\pdfoutput=1

\documentclass[11pt]{article}

\usepackage[]{EMNLP2023}

\usepackage{times}
\usepackage{latexsym}
\usepackage{amsmath, amsthm, amssymb, graphicx}
\usepackage{CJKutf8}
\usepackage[T1]{fontenc}
\usepackage{booktabs}
\usepackage{diagbox}
\usepackage[utf8]{inputenc}
\usepackage{microtype}
\usepackage{amsmath, amssymb, mathtools}
\usepackage{inconsolata}
\usepackage{enumerate}
\usepackage{caption}
\usepackage{cleveref}

\crefname{table}{Table}{Tables}
\Crefname{table}{Table}{Tables}

\title{ZhuJiu: A Multi-dimensional, Multi-faceted Chinese Benchmark for Large Language Models}

\author{
    Baoli Zhang$^{1}$\footnotemark[1], Haining Xie$^{1,2}$\footnotemark[1], Pengfan Du$^{1,2}$, Junhao Chen$^{3}$, Pengfei Cao$^{1}$,\\ \textbf{Yubo Chen}$^{1,2}$, \textbf{Shengping Liu}$^{4}$, \textbf{Kang Liu}$^{1,2}$ and \textbf{Jun Zhao}$^{1,2}$ \\
  $^1$Institute of Automation, Chinese Academy of Sciences \\
  $^2$School of Artificial Intelligence, University of Chinese Academy of Sciences \\
  $^3$ Harbin Engineering University, $^4$  Beijing Unisound Information Technology Co., Ltd \\
  \texttt{\{baoli.zhang,pengfei.cao,yubo.chen,kliu,jzhao\}@nlpr.ia.ac.cn}, \texttt{yisuanwang@hrbeu.edu.cn}\\
  \texttt{\{xiehaining21,dupengfan22\}@mails.ucas.ac.cn, liushengping@unisound.com} \\
}

\begin{document}
\maketitle
\footnotetext[1]{*Co-first authors, they contributed equally to this work.\\}
\begin{abstract}
The unprecedented performance of large language models (LLMs) requires comprehensive and accurate evaluation. We argue that for LLMs evaluation, benchmarks need to be comprehensive and systematic. To this end, we propose the ZhuJiu benchmark, which has the following strengths: (1) \textbf{Multi-dimensional ability coverage:} We comprehensively evaluate LLMs across 7 ability dimensions covering 51 tasks. Especially, we also propose a new benchmark that focuses on knowledge ability of LLMs. (2) \textbf{Multi-faceted evaluation methods collaboration:} We use 3 different yet complementary evaluation methods to comprehensively evaluate LLMs, which can ensure the authority and accuracy of the evaluation results. (3) \textbf{Comprehensive Chinese benchmark:} ZhuJiu is the pioneering benchmark that fully assesses LLMs in Chinese, while also providing equally robust evaluation abilities in English. (4) \textbf{Avoiding potential data leakage: }To avoid data leakage, we construct evaluation data specifically for 37 tasks. We evaluate 10 current mainstream LLMs and conduct an in-depth discussion and analysis of their results. The ZhuJiu benchmark and open-participation leaderboard are publicly released at \url{http://www.zhujiu-benchmark.com/} and we also provide a demo video at \url{https://youtu.be/qypkJ89L1Ic}.
\end{abstract}

\section{Introduction}

With the continuous development of large language models (LLMs), the emergence of GPT4 \citep{openai2023gpt4} is enough to trigger a new wave of technology. Various types of LLMs have recently been rapidly developing, such as Llama2 \citep{touvron2023llama}  and ChatGLM2 \citep{du2022glm}, demonstrating impressive generalization abilities and broad applicability. Therefore, it is crucial to conduct comprehensive and objective evaluations of LLMs to fully understand their strengths and limitations.

\begin{figure*}[t]
\centering
\includegraphics[width=\linewidth]{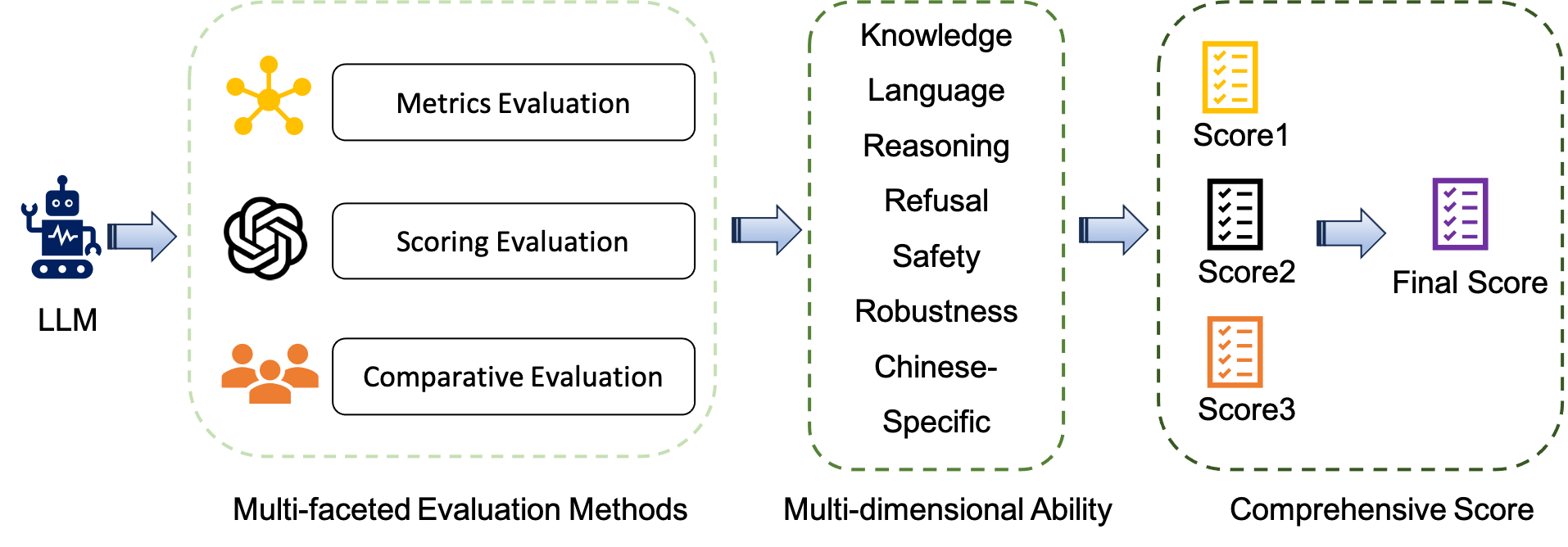}
\caption{The evaluation process of LLM using ZhuJiu.}
\label{fig:figure1}
\end{figure*}
 Specifically, on the one hand, for \textbf{applicators}, they need to understand the overall performance of LLMs or the advantages of LLMs in a specific aspect. Constructing comprehensive and authoritative benchmarks can help applicators significantly improve the efficiency of using LLMs. On the other hand, for \textbf{developers}, the improvement direction of LLMs requires accurate evaluation results as guidance.  An objective and fair benchmark can help them carry out relevant research work on LLMs more targetedly. 
 
To this end, scholars conduct extensive research on evaluations for LLMs and construct some superior benchmarks. 
Normally, the evaluation for LLMs includes two aspects: ability evaluation and evaluation method. Although \textbf{traditional benchmarks} such as GLUE \citep{wang2018glue}, SuperGLUE \citep{wang2019superglue} and CUGE \citep{yao2021cuge} still have a role to play in evaluating LLMs, their limitations are becoming increasingly apparent due to the growing diversity of evaluation dimensions and methods for LLMs.
For the \textbf{ability evaluation} of LLMs, recent work  proposes excellent benchmarks for LLMs in one or several aspects, such as knowledge, reasoning, language, safety and hallucination \citep{liang2022holistic,yu2023kola,sun2023safety,amayuelas2023knowledge,li2023halueval}. However, a comprehensive evaluation of LLMs remains insufficient.
For the \textbf{evaluation method} of LLMs, there are currently 3 main categories: (1) \textit{Metrics Evaluation}: Evaluating LLMs using existing datasets and corresponding metrics \citep{liang2022holistic}; (2) \textit{ChatGPT Evaluation}: Using GPT-like LLMs to generate evaluation data and  compare the response results of different LLMs \citep{pandalm2023}; (3) \textit{Model Arena}: constructing one-on-one model arenas where humans compare the evaluation results of models based on their own judgment \citep{zheng2023judging}.

Despite these successful efforts for LLMs' evaluations, existing studies still suffer from several limitations: (1) Current benchmarks tend to focus on evaluating LLMs on a single dimension of their abilities, which can not provide a comprehensive evaluation of LLMs. (2) Most benchmarks only use a single evaluation method, which may not provide an accurate evaluation of all the abilities of LLMs. For example, while HELM\citep{liang2022holistic} uses metrics to evaluate LLMs, it may not measure all abilities such as  long-text generation or machine translation, etc. (3) The cross-lingual abilities of LLMs, especially for Chinese, have garnered growing attention. However, the lack of a comprehensive Chinese benchmark for LLMs remains a critical issue. (4) Many current benchmarks only use public datasets for evaluation, risking potential data leakage. The results of evaluations based on this data lack credibility.

In this paper, we propose the ZhuJiu Benchmark to solve above mentioned problems, which can fill the gap in the development of a comprehensive benchmark for evaluating LLMs in Chinese. The advantages of the ZhuJiu are as follows: (1) \textbf{Multi-dimensional ability coverage:} we evaluate LLMs from 7 ability dimensions, including \textit{knowledge, Chinese-specific, language, reasoning, refusal, safety and robustness abilities}, covering 51 datasets to provide a comprehensive performance assessment. In addition, we also proposed a new paradigm for evaluating the knowledge ability. (2) \textbf{Multi-faceted evaluation methods coordination:} we use \textit{Metrics Evaluation}, \textit{Scoring Evaluation}, and \textit{Comparative Evaluation} for comprehensively evaluating LLMs  to ensure authoritative and accurate evaluation results. (3) \textbf{Comprehensive Chinese benchmark:} ZhuJiu is the pioneering Chinese benchmark that can comprehensively evaluate LLMs, while allowing equivalent assessment in English. (4) \textbf{Avoiding potential data leakage: } in addition to collecting 14 commonly used datasets, we construct 37  datasets for the evaluation of LLMs, ensuring maximum avoidance of data leakage and evaluation fairness. The overall evaluation process is shown in Figure \ref{fig:figure1}.

We also release an online evaluation platform that supports multiple functions including visualizations of evaluation results, participating in model arena and submission of evaluation model, etc.  Moreover, we  evaluate 10 publicly available LLMs, including ChatGLM \citep{du2022glm}, BELLE \citep{BELLE}, ChatGPT \citep{openai202chatgpt}, and so on. Based on the experimental results, we observe some interesting phenomena and summarize them in \ref{sec:performance}.

In summary, the contributions of this paper are as follows:
\begin{itemize}
  \item We propose ZhuJiu, the first Chinese benchmark that covers multi-dimensions of ability and employs multi-faceted evaluation methods in collaboration. Meanwhile in the ZhuJiu we  construct a novel benchmark for evaluating knowledge ability and 37 evaluation datasets to prevent data leakage issues.
  \item We  release an online evaluation platform that enables users to evaluate LLMs. We will continue to improve the platform, and update the evaluation leaderboard.
  \item Using the ZhuJiu benchmark, we evaluate 10 current LLMs, to comprehensively and deeply explore their abilities, providing valuable insights to inform future LLM development.
\begin{figure*}
    \centering
    \includegraphics[width=\linewidth]{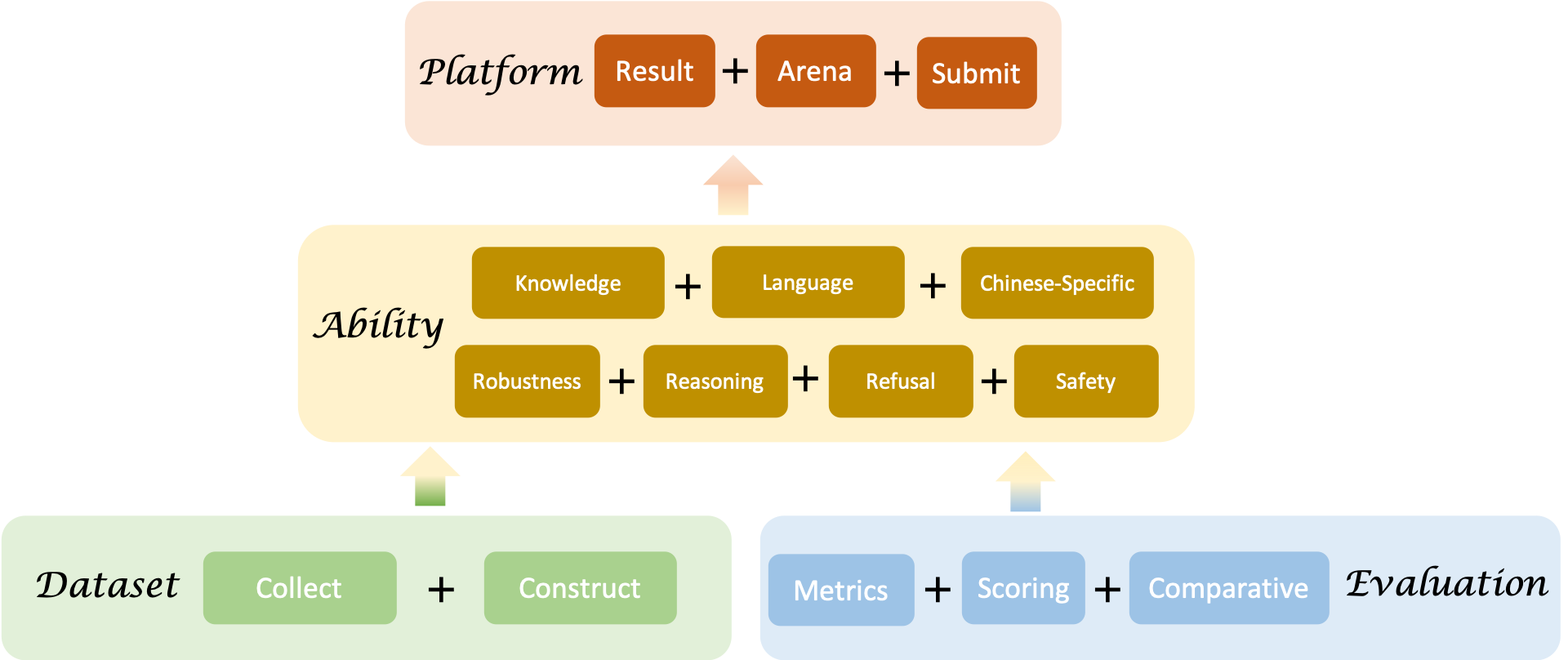}
    \caption{Overall view of the ZhuJiu benchmark. In ZhuJiu's framework, the integration of \textbf{multi-angle datasets} and \textbf{multi-faceted evaluation methods}  provides strong support for \textbf{multi-dimensional ability} assessment. Based on this, we have further developed an \textbf{online assessment platform} to support ZhuJiu's online assessment and result updates.}
    \label{fig:figure2}
\end{figure*}
  
\end{itemize}

\section{ZhuJiu Benchmark}
As stated above, the ZhuJiu benchmark uses 3 evaluation methods to assess the abilities across seven dimensions of LLMs.  This section provides a detailed introduction to the ZhuJiu benchmark covering the evaluation methods, datasets, and ability dimensions. We also detail the specific scoring rules in  Appendix \ref{sec:appendix scoring}. The evaluation framework is shown in Figure \ref{fig:figure2}.

\subsection{Evaluation Methods}
Unlike previous works that only use a single evaluation method \citep{liang2022holistic,PandaLM,pandalm2023,zheng2023judging}, in order to ensure the reliability of the evaluation results, we employ a collaborative evaluation approach that utilizes 3 types of evaluation methods: Metrics Evaluation, Scoring Evaluation, and Comparative Evaluation.

\subsubsection{Metrics Evaluation}
Metrics Evaluation is an indispensable component in LLM assessment, providing objective results \citep{chang2023survey}. In this paper, we adopt the HELM evaluation framework. Building on HELM \citep{liang2022holistic}, we extend it with additional Chinese benchmarks for language, reasoning, knowledge, and Chinese abilities, with 14 expanded datasets total.


\subsubsection{Scoring Evaluation}
The abilities demonstrated by ChatGPT \citep{openai202chatgpt} and GPT-4 \citep{openai2023gpt4} have brought us great surprises. Therefore, we conduct evaluations on the responses of LLMs using prompt engineering based on ChatGPT. Specifically, we evaluate different abilities and devise different perspectives to assist ChatGPT in scoring the responses. We use few-shot \citep{snell2017prototypical,ravi2016optimization,wang2020generalizing} method and answer label, combined with numerous experiments, to ensure the accuracy and stability of ChatGPT's evaluation results. 

\subsubsection{Comparative Evaluation}
Comparative evaluation is the most intuitive evaluation method. In this paper, we drew inspiration from the work of Chatbot Arena \citep{zheng2023judging} and used the \textit{one-on-one model arena method} to compare and evaluate the performance of LLMs based on human  judgments. Furthermore, we provide a one-on-one model comparison function in the platform, which allows users to compare the quality of responses from different LLMs to the same question.

\subsection{Datasets}
For a benchmark, the most crucial part is undoubtedly its data source and data quality. In ZhuJiu, our evaluation data comes from two parts. On the one hand, we use 14 currently popular LLMs evaluation datasets. On the other hand, considering the serious issue of data leakage when solely using public datasets for LLMs evaluation, which could compromise the fairness of evaluation results, we constructed 37 evaluation datasets based on ChatGPT \citep{openai202chatgpt}.

\subsubsection{Collect Datasets}
To ensure the generality of ZhuJiu, we evaluate LLMs using 14 publicly available datasets, which are essential due to their high quality and ability to accurately evaluate the performance of LLMs in certain aspects.

\subsubsection{Construct Datasets}
To address the issue of data leakage in LLMs evaluation, we are inspired by PandaLM \citep{pandalm2023} and we construct corresponding evaluation datasets for 37 specific tasks. Specifically, for each task, we first carefully select some evaluation data as seeds manually. Then, we use these seeds to generate prompts based on ChatGPT through self-instruction \citep{selfinstruct}. After that, we manually review and confirm the prompts we used (for each specific task, we generate 100 prompts in Chinese).

To better understand the processes of data construction and evaluation in a more intuitive way, we take Scoring Evaluation as an example to demonstrate the process, as shown in Figure \ref{fig:figure3}.



\subsection{Ability System}\label{sec:description}
With the help of the aforementioned evaluation methods and datasets, we can assess the abilities of LLMs in 7 aspects. We will provide a detailed introduction to the specific evaluation methods and details in this section.

\subsubsection{Knowledge Ability}
To comprehensively evaluate the knowledge abilities of LLMs, we conduct the evaluation from four perspectives: \textit{world knowledge}, \textit{commonsense knowledge}, \textit{linguistic knowledge}, and \textit{concept}. For each evaluation perspective, we select the appropriate properties of accuracy, robustness, completeness, and timeliness to construct evaluation datasets for evaluating LLMs. Detailed descriptions of these four properties are provided in Appendix \ref{sec:appendix knowledge}, using a detailed framework shown in Figure \ref{fig:figure4}. 
Compared to KoLA \citep{yu2023kola}, our evaluation  perspective for knowledge is broader.

For \textbf{world knowledge}, on the one hand, we utilize the GAOKAO-bench \citep{Zhang2023EvaluatingTP} (Non-mathematical section) and combine it with Metrics Evaluation to conduct the evaluation.  On the other hand, we construct corresponding evaluation datasets for each evaluation property, including accuracy, robustness, completeness, and timeliness, and evaluate LLMs using Scoring Evaluation.


For \textbf{commonsense knowledge}, we select commonsense triplets as the basic data and construct evaluation datasets based on the evaluation properties of accuracy and robustness. We then use Scoring Evaluation to evaluate LLMs.

For \textbf{linguistic  knowledge}, we use Chinese FrameNet (CFN) \citep{hao2007description,baker1998berkeley} as the original corpus. In order to simplify the evaluation form of linguistic knowledge, we mainly construct datasets in the following two ways: one is to infer the ``frame name'' of the linguistic frame according to the ``frame def'' in the linguistic frame, the other is to infer the ``frame name'' of the linguistic frame based on the ``lexical-unit name'' in the linguistic frame. Then we can  evaluate the accuracy and robustness of LLMs linguistic knowledge by using the Scoring Evaluation.

For \textbf{concept}, we manually select common entity words as the original data and evaluate the accuracy and robustness of LLMs concepts with Scoring Evaluation.

\subsubsection{Chinese-Specific Ability}
Following SuperCLUE \citep{SuperCLUE}, and  conventional Chinese evaluations, the Chinese-specific ability evaluation aims to use corpora with Chinese unique characteristics as the original data to form evaluation data. These corpora include ChID \citep{zheng-etal-2019-chid}, CCPM \citep{li2021CCPM}, CINLID and YACLC \citep{wang2021yaclc}, and we evaluate LLMs using Metrics Evaluation.
\subsubsection{Language Ability}
We conduct a comprehensive evaluation of LLMs' language ability from both aspects of language understanding and language generation. For evaluating LLMs' \textbf{language understanding ability}, we choose to evaluate them on the tasks of reading comprehension and coreference resolution. We find that using existing datasets could achieve good evaluation results, and the datasets we use included C3 \citep{sun2020investigating}, GCRC \citep{tan2021gcrc}, CMRC \citep{cui2018span}, DRCR \citep{shao2018drcd} and CLUEWSC-2020 \citep{xu-etal-2020-clue}, correspondingly  we use Metrics Evaluation. For evaluating LLMs' \textbf{language generation ability}, we summarize 6 typical language generation tasks, including \textit{common response} (Daily question answering), \textit{dialogue} (Dialog generation based on the scene), \textit{formal writing} (Generation of formal texts for letters and other formal occasions), \textit{poetry} (Generate poems on request), \textit{writing story} (Generate stories on request) and \textit{writing style} (Generate text according to the requirements of the writing style) \citep{chang2023survey}, and evaluating by Scoring Evaluation. 

\subsubsection{Reasoning Ability}
As the evaluation of LLMs' reasoning ability is less affected by data leakage \citep{chang2023survey}, we find that only using publicly available datasets could yield relatively fair results. We select the currently popular mathematical reasoning and text semantic reasoning tasks, and the datasets included GAOKAO-bench \citep{Zhang2023EvaluatingTP} (mathematics section), Math23k \citep{wang2017deep}, OCNLI \citep{hu2020ocnli}, Chinese-SNLI \citep{chinese-snli} and Chinese-MNLI \citep{xu-etal-2020-clue}. The evaluation method for reasoning ability is based on Metrics Evaluation.

\subsubsection{Refusal Ability}
Regarding the refusal ability, we can understand it like this: \textit{To know what you know and to know what you do not know, that is true knowledge}. For constructing datasets of refusal ability, we drew inspiration from the categories of Known-Unknown Questions proposed in \citealp{amayuelas2023knowledge}, including \textit{Future Unknown}, \textit{Unsolved Problem/Mystery}, \textit{Controversial/Debatable Question}, \textit{Question with False Assumption}, \textit{Counterfactual Question} and \textit{Underspecified Question}. Then, we employ Scoring Evaluation to assess LLMs for each category.

\subsubsection{Safety}
For the evaluation of safety ability, we following \citealp{sun2023safety}'s classification of safety ability and further summarize and categorize them. We derive a total of 9 evaluation tasks from 6 perspectives, including \textit{Insult}, \textit{Human Health (Physical harm and Mental health)}, \textit{Social Topic (Unfairness discrimination and Ethics morality)}, \textit{Serious Risk (Criminal Activity and Unsafe Instruction Topic)}, \textit{Goal Hijacking} and \textit{Role play instruction}. Subsequently, we employ the Scoring Evaluation to assess LLMs.

\subsubsection{Robustness}
Traditional robustness evaluation primarily focuses on assessing the impact of adding perturbations of varying granularity to the text on the performance of the model \citep{zhu2023promptbench,wang-etal-2021-textflint,wang2023robustness}. Regarding the robustness evaluation of LLMs, on one hand, we still consider token-level perturbations and sentence-level perturbations from the traditional robustness evaluation perspective, and propose three evaluation tasks including \textit{Error Message}, \textit{Redundant Information} and \textit{Redundant Dialogue}. On the other hand, we expand three aspects of \textit{Format Output}, \textit{Dialect} and \textit{Unique Solution tasks} (Evaluate the certainty of the model's answer to the unique solution through multiple rounds of questioning) specifically tailor to the characteristics of LLMs. Ultimately, we conduct evaluations on these six aspects based on the Scoring Evaluation.

\begin{table*}[t]
    \centering
    \resizebox{\linewidth}{!}{
        \begin{tabular}{c | c c c c c c c | c}
        \toprule
        \diagbox{LLMs}{Score}{Abilities} & Knowledge & Chinese-Specific & Language & Reasoning & Refusal & Safety & Robustness & All  \\
        \midrule
        ChatGLM2-6B  & \textbf{91.1} & 59.5 & \textbf{85.6} & \textbf{80.6} & \textbf{82.0} & 55.4 & \textbf{63.8} & \textbf{74.0} \\ 
        ChatGLM-6B   & 67.3 & \textbf{73.9} & 74.8 & 37.0 & 80.4 & \textbf{82.3} & 50.0 & 66.5 \\
        BELLE-7B     & 54.53 & 40.54 & 54.2 & 44.5 & 58.1 & 39.8 & 55.9 & 49.6 \\
        Moss-Moon-003-SFT   & 50.4 & 27.0 & 56.3 & 15.9 & 48.2 & 64.8 & 46.2 & 44.1 \\
        ChatYuan-large-v2   & 58.8 & 20.7 & 37.3 & 42.7 & 37.5 & 78.1 & 29.8 & 43.6 \\
        ChatFlow     & 43.3 &  54.1  & 33.3 &  47.1  & 39.2 & 40.3 & 36.1 & 41.9 \\
        Phoenix-Inst-chat-7B     & 19.53 & 0 & 62.3 & 0 & 67.3 & 65.9 & 61.0 & 39.4 \\
        RWKV         & 23.4 &  15.0 & 35.8  &  69.3  & 16.4 & 20.5 & 45.9 & 32.3\\
        Baichuan-7B  & 34.6 & 41.4 & 19.7 & 43.6 & 0   &  0  & 32.4 & 24.5 \\

        \hline
        GPT-3.5-turbo   & 82.4 & 100.0 & 84.3 & 100.0 & 100.0 & 100.0 & 85.5 & 93.2 \\
        \bottomrule
        
        \end{tabular}
        
    }
    \caption{The overall performance based on ten-point system of the LLMs participating in the ZhuJiu evaluation in the first season. The score of GPT-3.5-turbo is only for reference and not included in the evaluation.}
    \label{Tab:Table1}
\end{table*}


\section{Platform}
We develop an online platform  to provide a range of services for the community as follows: 

\textbf{Visualizations of evaluation results}  We  publish the rankings of all model evaluations on the platform, including specific scores for each ability and evaluation method, and the rankings will be updated continuously as the evaluations progress. The visualization result webpage is shown in Figure \ref{fig:figure5}.

\textbf{Participating in Model Arena} We launch an one-on-one model arena feature on our platform, where everyone can support the LLMs they believe perform better based on their own judgment. Please refer to Figure \ref{fig:figure6} to see the web view of the model arena.

\textbf{Submission of Evaluation Model} We also encourage everyone to actively participate in our evaluations and join the leaderboard. On our platform, we allow users to submit applications for evaluations.

\section{Experiment}
\subsection{Evaluated Models}
To facilitate the utilization and advancement of LLMs, the primary emphasis of ZhuJiu's inaugural evaluation phase is directed towards \textit{open-source} LLMs with a parameter magnitude of approximately 10 billion, including: ChatGLM-6B \citep{du2022glm}, ChatGLM2-6B \citep{du2022glm}, BELLE-7B \citep{BELLE}, Baichuan-7B \citep{2023baichuan7b}, ChatFlow \citep{li-etal-2022-csl,zhao2022tencentpretrain}, Phoenix-Inst-Chat-7B \citep{phoenix-2023,llm-zoo-2023}, ChatYuan-large-v2 \citep{clueai2023chatyuan}, Moss-Moon-003-SFT \citep{sun2023moss} and RWKV \citep{peng_bo_2021_5196578}. Concurrently, we employ ChatGPT \citep{openai202chatgpt} as a comparative benchmark and conduct an assessment of the GPT-3.5-turbo \textit{API service}.

\subsection{Overall Performance}\label{sec:performance}
We report the overall performance in Table \ref{Tab:Table1}, and show more detailed assessment results in our platform. From the results, we can obtain some intriguing findings:
\begin{enumerate}[(1)]
    \item \textbf{Model-size Determines Performance:} Based on the results in table \ref{Tab:Table1}, it becomes evident that models with a parameter size of around 10 billion still exhibit significant limitations in overall performance compared to GPT-3.5-turbo \citep{openai202chatgpt}. In ZhuJiu, the performance of most LLMs is relatively mediocre, with ChatGLM2 and ChatGLM \citep{du2022glm} showing relatively better performance. It becomes apparent that the size of the model's parameters continues to play a vital role in determining its performance.
    \item \textbf{Lower Limit Sets Upper Limit:} The analysis reveals that Phoenix \citep{phoenix-2023} demonstrates notable proficiency in refusal and safety abilities, etc. However, its overall ranking is comparatively lower, primarily attributed to its limitations in reasoning and Chinese-specific abilities. These deficiencies are also observed in other LLMs occupying lower positions in the rankings. However, \textit{the lower limits of various abilities in LLMs often determine the upper limits of LLMs' application prospects}.
    \item  \textbf{Knowledge is Power:}  In ZhuJiu, our primary focus lies in the knowledge ability of LLMs, as the pivotal task at hand is to ensure LLMs acquire accurate knowledge and effectively harness their acquired knowledge. However, in this season, the majority of LLMs  exhibit subpar performance in terms of knowledge capacity, making the ZhuJiu benchmark exceptionally challenging. 
    The results reveal that ChatGLM2 \citep{du2022glm} exhibits strong performance in knowledge ability, surpassing even ChatGPT. This can be attributed to ChatGLM2 leveraging a larger and higher-quality Chinese training corpus. 
\end{enumerate}

\section{Conclusion and Future Work}
In this work, we present ZhuJiu, the pioneering multi-dimensional ability coverage, multi-faceted evaluation methods collaboration Chinese benchmark. ZhuJiu is capable of using 3 evaluation methods to comprehensively evaluate LLMs across 7 ability dimensions, using 51 datasets. Additionally, we independently construct 37 evaluation datasets to maximize the avoidance of data leakage issues in LLM evaluation. We also focus on expanding the evaluation of knowledge ability, providing a new framework for assessing LLMs' knowledge ability. Finally, we provide a comprehensive and continuously updated evaluation platform with multiple functions and in the first season of ZhuJiu, we evaluate 10 \textit{open-source} LLMs.


In the future, we plan to (1) continuously construct high-quality evaluation datasets to enrich ZhuJiu, (2) further perfect the assessment of knowledge ability and develop new evaluation methods for Chinese characteristic ability, (3) further perfect the platform's functionality and update the platform's information.

\bibliography{anthology,custom}

\begin{thebibliography}{47}
\expandafter\ifx\csname natexlab\endcsname\relax\def\natexlab#1{#1}\fi

\bibitem[{chi(2019)}]{chinese-snli}
 2019.
\newblock Blog: Chinese-snli.
\newblock \url{https://gitee.com/jiaodaxin/CNSD}.

\bibitem[{202(2023)}]{2023baichuan7b}
 2023.
\newblock Blog: Baichuan-7b.
\newblock \url{https://github.com/baichuan-inc/Baichuan-7B}.

\bibitem[{Amayuelas et~al.(2023)Amayuelas, Pan, Chen, and
  Wang}]{amayuelas2023knowledge}
Alfonso Amayuelas, Liangming Pan, Wenhu Chen, and William Wang. 2023.
\newblock Knowledge of knowledge: Exploring known-unknowns uncertainty with
  large language models.
\newblock \emph{arXiv preprint arXiv:2305.13712}.

\bibitem[{Baker et~al.(1998)Baker, Fillmore, and Lowe}]{baker1998berkeley}
Collin~F Baker, Charles~J Fillmore, and John~B Lowe. 1998.
\newblock The berkeley framenet project.
\newblock In \emph{COLING 1998 Volume 1: The 17th International Conference on
  Computational Linguistics}.

\bibitem[{Bo(2021)}]{peng_bo_2021_5196578}
PENG Bo. 2021.
\newblock \href {https://doi.org/10.5281/zenodo.5196577} {Blinkdl/rwkv-lm:
  0.01}.

\bibitem[{Chang et~al.(2023)Chang, Wang, Wang, Wu, Zhu, Chen, Yang, Yi, Wang,
  Wang et~al.}]{chang2023survey}
Yupeng Chang, Xu~Wang, Jindong Wang, Yuan Wu, Kaijie Zhu, Hao Chen, Linyi Yang,
  Xiaoyuan Yi, Cunxiang Wang, Yidong Wang, et~al. 2023.
\newblock A survey on evaluation of large language models.
\newblock \emph{arXiv preprint arXiv:2307.03109}.

\bibitem[{Chen et~al.(2023{\natexlab{a}})Chen, Chen, Zhang, Jiang, Chen, Yu,
  Wang, Liang, Zhang, Zhang, Li, Wan, Li, and Wang}]{llm-zoo-2023}
Zhihong Chen, Junying Chen, Hongbo Zhang, Feng Jiang, Guiming Chen, Fei Yu,
  Tiannan Wang, Juhao Liang, Chen Zhang, Zhiyi Zhang, Jianquan Li, Xiang Wan,
  Haizhou Li, and Benyou Wang. 2023{\natexlab{a}}.
\newblock Llm zoo: democratizing chatgpt.
\newblock \url{https://github.com/FreedomIntelligence/LLMZoo}.

\bibitem[{Chen et~al.(2023{\natexlab{b}})Chen, Jiang, Chen, Wang, Yu, Chen,
  Zhang, Liang, Zhang, Zhang, Li, Wan, Wang, and Li}]{phoenix-2023}
Zhihong Chen, Feng Jiang, Junying Chen, Tiannan Wang, Fei Yu, Guiming Chen,
  Hongbo Zhang, Juhao Liang, Chen Zhang, Zhiyi Zhang, Jianquan Li, Xiang Wan,
  Benyou Wang, and Haizhou Li. 2023{\natexlab{b}}.
\newblock Phoenix: Democratizing chatgpt across languages.
\newblock \emph{arXiv preprint arXiv:2304.10453}.

\bibitem[{Cui et~al.(2018)Cui, Liu, Che, Xiao, Chen, Ma, Wang, and
  Hu}]{cui2018span}
Yiming Cui, Ting Liu, Wanxiang Che, Li~Xiao, Zhipeng Chen, Wentao Ma, Shijin
  Wang, and Guoping Hu. 2018.
\newblock A span-extraction dataset for chinese machine reading comprehension.
\newblock \emph{arXiv preprint arXiv:1810.07366}.

\bibitem[{Du et~al.(2022)Du, Qian, Liu, Ding, Qiu, Yang, and Tang}]{du2022glm}
Zhengxiao Du, Yujie Qian, Xiao Liu, Ming Ding, Jiezhong Qiu, Zhilin Yang, and
  Jie Tang. 2022.
\newblock Glm: General language model pretraining with autoregressive blank
  infilling.
\newblock In \emph{Proceedings of the 60th Annual Meeting of the Association
  for Computational Linguistics (Volume 1: Long Papers)}, pages 320--335.

\bibitem[{Hao et~al.(2007)Hao, Liu, Li, and Liu}]{hao2007description}
Xiaoyan Hao, Wei Liu, Ru~Li, and Kaiying Liu. 2007.
\newblock Description systems of the chinese framenet database and software
  tools.
\newblock \emph{Journal of Chinese information processing}, 21(5):96--100.

\bibitem[{Hu et~al.(2020)Hu, Richardson, Xu, Li, K{\"u}bler, and
  Moss}]{hu2020ocnli}
Hai Hu, Kyle Richardson, Liang Xu, Lu~Li, Sandra K{\"u}bler, and Lawrence~S
  Moss. 2020.
\newblock Ocnli: Original chinese natural language inference.
\newblock \emph{arXiv preprint arXiv:2010.05444}.

\bibitem[{Jifan~Yu(2023)}]{yu2023kola}
Shangqing Tu Shulin Cao Daniel Zhang-Li Xin Lv Hao Peng Zijun Yao Xiaohan Zhang
  Hanming Li Chunyang Li Zheyuan Zhang Yushi Bai Yantao Liu Amy Xin Nianyi Lin
  Kaifeng Yun Linlu Gong Jianhui Chen Zhili Wu Yunjia Qi Weikai Li Yong Guan
  Kaisheng Zeng Ji Qi Hailong Jin Jinxin Liu Yu Gu Yuan Yao Ning Ding Lei Hou
  Zhiyuan Liu Bin Xu Jie Tang Juanzi~Li Jifan~Yu, Xiaozhi~Wang. 2023.
\newblock \href {"https://github.com/THU-KEG/KoLA"} {Kola: Carefully
  benchmarking world knowledge of large language models}.

\bibitem[{Li et~al.(2023)Li, Cheng, Zhao, Nie, and Wen}]{li2023halueval}
Junyi Li, Xiaoxue Cheng, Wayne~Xin Zhao, Jian-Yun Nie, and Ji-Rong Wen. 2023.
\newblock Halueval: A large-scale hallucination evaluation benchmark for large
  language models.
\newblock \emph{arXiv e-prints}, pages arXiv--2305.

\bibitem[{Li et~al.(2021)Li, Qi, Sun, Yi, and Zhang}]{li2021CCPM}
Wenhao Li, Fanchao Qi, Maosong Sun, Xiaoyuan Yi, and Jiarui Zhang. 2021.
\newblock Ccpm: A chinese classical poetry matching dataset.
\newblock \emph{arXiv preprint arXiv:2106.01979}.

\bibitem[{Li et~al.(2022)Li, Zhang, Zhao, Shen, Liu, Mao, and
  Zhang}]{li-etal-2022-csl}
Yudong Li, Yuqing Zhang, Zhe Zhao, Linlin Shen, Weijie Liu, Weiquan Mao, and
  Hui Zhang. 2022.
\newblock \href {https://aclanthology.org/2022.coling-1.344} {{CSL}: A
  large-scale {C}hinese scientific literature dataset}.
\newblock In \emph{Proceedings of the 29th International Conference on
  Computational Linguistics}, pages 3917--3923, Gyeongju, Republic of Korea.
  International Committee on Computational Linguistics.

\bibitem[{Liang et~al.(2022)Liang, Bommasani, Lee, Tsipras, Soylu, Yasunaga,
  Zhang, Narayanan, Wu, Kumar et~al.}]{liang2022holistic}
Percy Liang, Rishi Bommasani, Tony Lee, Dimitris Tsipras, Dilara Soylu,
  Michihiro Yasunaga, Yian Zhang, Deepak Narayanan, Yuhuai Wu, Ananya Kumar,
  et~al. 2022.
\newblock Holistic evaluation of language models.
\newblock \emph{arXiv preprint arXiv:2211.09110}.

\bibitem[{Liang~Xu and others~from SuperCLUE~team(2023)}]{SuperCLUE}
Kangkang Zhao Lei~Zhu Liang~Xu, Xuanwei~Zhang and others~from SuperCLUE~team.
  2023.
\newblock Superclue: A benchmark for foundation models in chinese.
\newblock \url{https://github.com/CLUEbench/SuperCLUE}.

\bibitem[{OpenAI(2022)}]{openai202chatgpt}
OpenAI. 2022.
\newblock Blog: Introducing chatgpt.
\newblock \url{https://openai.com/blog/chatgpt}.

\bibitem[{OpenAI(2023)}]{openai2023gpt4}
OpenAI. 2023.
\newblock \href {http://arxiv.org/abs/2303.08774} {Gpt-4 technical report}.

\bibitem[{Ravi and Larochelle(2016)}]{ravi2016optimization}
Sachin Ravi and Hugo Larochelle. 2016.
\newblock Optimization as a model for few-shot learning.
\newblock In \emph{International conference on learning representations}.

\bibitem[{Shao et~al.(2018)Shao, Liu, Lai, Tseng, and Tsai}]{shao2018drcd}
Chih~Chieh Shao, Trois Liu, Yuting Lai, Yiying Tseng, and Sam Tsai. 2018.
\newblock Drcd: A chinese machine reading comprehension dataset.
\newblock \emph{arXiv preprint arXiv:1806.00920}.

\bibitem[{Snell et~al.(2017)Snell, Swersky, and Zemel}]{snell2017prototypical}
Jake Snell, Kevin Swersky, and Richard Zemel. 2017.
\newblock Prototypical networks for few-shot learning.
\newblock \emph{Advances in neural information processing systems}, 30.

\bibitem[{Sun et~al.(2023{\natexlab{a}})Sun, Zhang, Deng, Cheng, and
  Huang}]{sun2023safety}
Hao Sun, Zhexin Zhang, Jiawen Deng, Jiale Cheng, and Minlie Huang.
  2023{\natexlab{a}}.
\newblock Safety assessment of chinese large language models.
\newblock \emph{arXiv preprint arXiv:2304.10436}.

\bibitem[{Sun et~al.(2020)Sun, Yu, Yu, and Cardie}]{sun2020investigating}
Kai Sun, Dian Yu, Dong Yu, and Claire Cardie. 2020.
\newblock Investigating prior knowledge for challenging chinese machine reading
  comprehension.
\newblock \emph{Transactions of the Association for Computational Linguistics},
  8:141--155.

\bibitem[{Sun et~al.(2023{\natexlab{b}})Sun, Zhang, He, Li, Cheng, Yan, Liu,
  Shao, Tang, Zhao, Chen, Zheng, Zhou, Li, Zhan, Zhou, Li, Yang, Wu, Yin,
  Huang, and Qiu}]{sun2023moss}
Tianxiang Sun, Xiaotian Zhang, Zhengfu He, Peng Li, Qinyuan Cheng, Hang Yan,
  Xiangyang Liu, Yunfan Shao, Qiong Tang, Xingjian Zhao, Ke~Chen, Yining Zheng,
  Zhejian Zhou, Ruixiao Li, Jun Zhan, Yunhua Zhou, Linyang Li, Xiaogui Yang,
  Lingling Wu, Zhangyue Yin, Xuanjing Huang, and Xipeng Qiu.
  2023{\natexlab{b}}.
\newblock Moss: Training conversational language models from synthetic data.

\bibitem[{Tan et~al.(2021)Tan, Wang, Ji, Li, Li, Hu, Zhao, and
  Han}]{tan2021gcrc}
Hongye Tan, Xiaoyue Wang, Yu~Ji, Ru~Li, Xiaoli Li, Zhiwei Hu, Yunxiao Zhao, and
  Xiaoqi Han. 2021.
\newblock Gcrc: A new challenging mrc dataset from gaokao chinese for
  explainable evaluation.
\newblock In \emph{Findings of the Association for Computational Linguistics:
  ACL-IJCNLP 2021}, pages 1319--1330.

\bibitem[{Touvron et~al.(2023)Touvron, Martin, Stone, Albert, Almahairi,
  Babaei, Bashlykov, Batra, Bhargava, Bhosale et~al.}]{touvron2023llama}
Hugo Touvron, Louis Martin, Kevin Stone, Peter Albert, Amjad Almahairi, Yasmine
  Babaei, Nikolay Bashlykov, Soumya Batra, Prajjwal Bhargava, Shruti Bhosale,
  et~al. 2023.
\newblock Llama 2: Open foundation and fine-tuned chat models.
\newblock \emph{arXiv preprint arXiv:2307.09288}.

\bibitem[{Wang et~al.(2019)Wang, Pruksachatkun, Nangia, Singh, Michael, Hill,
  Levy, and Bowman}]{wang2019superglue}
Alex Wang, Yada Pruksachatkun, Nikita Nangia, Amanpreet Singh, Julian Michael,
  Felix Hill, Omer Levy, and Samuel Bowman. 2019.
\newblock Superglue: A stickier benchmark for general-purpose language
  understanding systems.
\newblock \emph{Advances in neural information processing systems}, 32.

\bibitem[{Wang et~al.(2018)Wang, Singh, Michael, Hill, Levy, and
  Bowman}]{wang2018glue}
Alex Wang, Amanpreet Singh, Julian Michael, Felix Hill, Omer Levy, and Samuel~R
  Bowman. 2018.
\newblock Glue: A multi-task benchmark and analysis platform for natural
  language understanding.
\newblock \emph{arXiv preprint arXiv:1804.07461}.

\bibitem[{Wang et~al.(2023{\natexlab{a}})Wang, Hu, Hou, Chen, Zheng, Wang,
  Yang, Huang, Ye, Geng et~al.}]{wang2023robustness}
Jindong Wang, Xixu Hu, Wenxin Hou, Hao Chen, Runkai Zheng, Yidong Wang, Linyi
  Yang, Haojun Huang, Wei Ye, Xiubo Geng, et~al. 2023{\natexlab{a}}.
\newblock On the robustness of chatgpt: An adversarial and out-of-distribution
  perspective.
\newblock \emph{arXiv preprint arXiv:2302.12095}.

\bibitem[{Wang et~al.(2021{\natexlab{a}})Wang, Liu, Gui, Zhang
  et~al.}]{wang-etal-2021-textflint}
Xiao Wang, Qin Liu, Tao Gui, Qi~Zhang, et~al. 2021{\natexlab{a}}.
\newblock \href {https://doi.org/10.18653/v1/2021.acl-demo.41} {Textflint:
  Unified multilingual robustness evaluation toolkit for natural language
  processing}.
\newblock In \emph{Proceedings of the 59th Annual Meeting of the Association
  for Computational Linguistics and the 11th International Joint Conference on
  Natural Language Processing: System Demonstrations}, pages 347--355, Online.
  Association for Computational Linguistics.

\bibitem[{Wang et~al.(2017)Wang, Liu, and Shi}]{wang2017deep}
Yan Wang, Xiaojiang Liu, and Shuming Shi. 2017.
\newblock Deep neural solver for math word problems.
\newblock In \emph{Proceedings of the 2017 conference on empirical methods in
  natural language processing}, pages 845--854.

\bibitem[{Wang et~al.(2020)Wang, Yao, Kwok, and Ni}]{wang2020generalizing}
Yaqing Wang, Quanming Yao, James~T Kwok, and Lionel~M Ni. 2020.
\newblock Generalizing from a few examples: A survey on few-shot learning.
\newblock \emph{ACM computing surveys (csur)}, 53(3):1--34.

\bibitem[{Wang et~al.(2023{\natexlab{b}})Wang, Yu, Zeng, Yang, Heng, Wang,
  Chen, Jiang, Xie, Wang, Xie, Ye, Zhang, and Zhang}]{PandaLM}
Yidong Wang, Zhuohao Yu, Zhengran Zeng, Linyi Yang, Qiang Heng, Cunxiang Wang,
  Hao Chen, Chaoya Jiang, Rui Xie, Jindong Wang, Xing Xie, Wei Ye, Shikun
  Zhang, and Yue Zhang. 2023{\natexlab{b}}.
\newblock Pandalm: Reproducible and automated language model assessment.
\newblock \url{https://github.com/WeOpenML/PandaLM}.

\bibitem[{Wang et~al.(2023{\natexlab{c}})Wang, Yu, Zeng, Yang, Wang, Chen,
  Jiang, Xie, Wang, Xie, Ye, Zhang, and Zhang}]{pandalm2023}
Yidong Wang, Zhuohao Yu, Zhengran Zeng, Linyi Yang, Cunxiang Wang, Hao Chen,
  Chaoya Jiang, Rui Xie, Jindong Wang, Xing Xie, Wei Ye, Shikun Zhang, and Yue
  Zhang. 2023{\natexlab{c}}.
\newblock Pandalm: An automatic evaluation benchmark for llm instruction tuning
  optimization.

\bibitem[{Wang et~al.(2021{\natexlab{b}})Wang, Kong, Yang, Wang, Lu, Hu, He,
  Liu, Chen, Yang et~al.}]{wang2021yaclc}
Yingying Wang, Cunliang Kong, Liner Yang, Yijun Wang, Xiaorong Lu, Renfen Hu,
  Shan He, Zhenghao Liu, Yun Chen, Erhong Yang, et~al. 2021{\natexlab{b}}.
\newblock Yaclc: A chinese learner corpus with multidimensional annotation.
\newblock \emph{arXiv preprint arXiv:2112.15043}.

\bibitem[{Wang et~al.(2022)Wang, Kordi, Mishra, Liu, Smith, Khashabi, and
  Hajishirzi}]{selfinstruct}
Yizhong Wang, Yeganeh Kordi, Swaroop Mishra, Alisa Liu, Noah~A. Smith, Daniel
  Khashabi, and Hannaneh Hajishirzi. 2022.
\newblock Self-instruct: Aligning language model with self generated
  instructions.

\bibitem[{Xu et~al.(2020)Xu, Hu, Zhang, Li, Cao, Li, Xu, Sun, Yu, Yu, Tian,
  Dong, Liu, Shi, Cui, Li, Zeng, Wang, Xie, Li, Patterson, Tian, Zhang, Zhou,
  Liu, Zhao, Zhao, Yue, Zhang, Yang, Richardson, and Lan}]{xu-etal-2020-clue}
Liang Xu, Hai Hu, Xuanwei Zhang, Lu~Li, Chenjie Cao, Yudong Li, Yechen Xu, Kai
  Sun, Dian Yu, Cong Yu, Yin Tian, Qianqian Dong, Weitang Liu, Bo~Shi, Yiming
  Cui, Junyi Li, Jun Zeng, Rongzhao Wang, Weijian Xie, Yanting Li, Yina
  Patterson, Zuoyu Tian, Yiwen Zhang, He~Zhou, Shaoweihua Liu, Zhe Zhao, Qipeng
  Zhao, Cong Yue, Xinrui Zhang, Zhengliang Yang, Kyle Richardson, and Zhenzhong
  Lan. 2020.
\newblock \href {https://doi.org/10.18653/v1/2020.coling-main.419} {{CLUE}: A
  {C}hinese language understanding evaluation benchmark}.
\newblock In \emph{Proceedings of the 28th International Conference on
  Computational Linguistics}, pages 4762--4772, Barcelona, Spain (Online).
  International Committee on Computational Linguistics.

\bibitem[{Xuanwei~Zhang and Zhao(2022)}]{clueai2023chatyuan}
Liang~Xu Xuanwei~Zhang and Kangkang Zhao. 2022.
\newblock \href {https://github.com/clue-ai/ChatYuan} {Chatyuan: A large
  language model for dialogue in chinese and english}.

\bibitem[{Yao et~al.(2021)Yao, Dong, Guan, Cao, Zhang, Xiao, Wang, Qi, Bao, Nie
  et~al.}]{yao2021cuge}
Yuan Yao, Qingxiu Dong, Jian Guan, Boxi Cao, Zhengyan Zhang, Chaojun Xiao,
  Xiaozhi Wang, Fanchao Qi, Junwei Bao, Jinran Nie, et~al. 2021.
\newblock Cuge: A chinese language understanding and generation evaluation
  benchmark.
\newblock \emph{arXiv preprint arXiv:2112.13610}.

\bibitem[{Yunjie~Ji and Li(2023)}]{BELLE}
Yan Gong Yiping Peng Qiang Niu-Baochang~Ma Yunjie~Ji, Yong~Deng and Xiangang
  Li. 2023.
\newblock Belle: Be everyone's large language model engine.
\newblock \url{https://github.com/LianjiaTech/BELLE}.

\bibitem[{Zhang et~al.(2023)Zhang, Li, Zong, Ying, He, and
  Qiu}]{Zhang2023EvaluatingTP}
Xiaotian Zhang, Chunyang Li, Yi~Zong, Zhengyu Ying, Liang He, and Xipeng Qiu.
  2023.
\newblock Evaluating the performance of large language models on gaokao
  benchmark.

\bibitem[{Zhao et~al.(2022)Zhao, Li, Hou, Zhao, Tian, Liu, Chen, Sun, Liu, Mao
  et~al.}]{zhao2022tencentpretrain}
Zhe Zhao, Yudong Li, Cheng Hou, Jing Zhao, Rong Tian, Weijie Liu, Yiren Chen,
  Ningyuan Sun, Haoyan Liu, Weiquan Mao, et~al. 2022.
\newblock Tencentpretrain: A scalable and flexible toolkit for pre-training
  models of different modalities.
\newblock \emph{arXiv preprint arXiv:2212.06385}.

\bibitem[{Zheng et~al.(2019)Zheng, Huang, and Sun}]{zheng-etal-2019-chid}
Chujie Zheng, Minlie Huang, and Aixin Sun. 2019.
\newblock {C}h{ID}: A large-scale {C}hinese {ID}iom dataset for cloze test.
\newblock In \emph{ACL}.

\bibitem[{Zheng et~al.(2023)Zheng, Chiang, Sheng, Zhuang, Wu, Zhuang, Lin, Li,
  Li, Xing, Zhang, Gonzalez, and Stoica}]{zheng2023judging}
Lianmin Zheng, Wei-Lin Chiang, Ying Sheng, Siyuan Zhuang, Zhanghao Wu, Yonghao
  Zhuang, Zi~Lin, Zhuohan Li, Dacheng Li, Eric.~P Xing, Hao Zhang, Joseph~E.
  Gonzalez, and Ion Stoica. 2023.
\newblock \href {http://arxiv.org/abs/2306.05685} {Judging llm-as-a-judge with
  mt-bench and chatbot arena}.

\bibitem[{Zhu et~al.(2023)Zhu, Wang, Zhou, Wang, Chen, Wang, Yang, Ye, Gong,
  Zhang et~al.}]{zhu2023promptbench}
Kaijie Zhu, Jindong Wang, Jiaheng Zhou, Zichen Wang, Hao Chen, Yidong Wang,
  Linyi Yang, Wei Ye, Neil~Zhenqiang Gong, Yue Zhang, et~al. 2023.
\newblock Promptbench: Towards evaluating the robustness of large language
  models on adversarial prompts.
\newblock \emph{arXiv preprint arXiv:2306.04528}.

\end{thebibliography}
\bibliographystyle{acl_natbib}

\appendix

\section{Scoring Rules}
\label{sec:appendix scoring}
We will comprehensively evaluate the model from seven ability dimensions and 3 assessment methods to ensure the thoroughness and authority of the evaluation results.Specifically, the comprehensive evaluation process can be broken down into three steps.

\textbf{Step 1} For each ability dimension score $A$, we will take the average of LLM's scores $ \textbf{d} = [d_{1}, \ldots, d_{n}] $ on each dataset as LLM's score for that ability dimension:
\begin{equation}
A = \frac{1}{n} \sum_{i=1}^{n} d_i  
\end{equation}

\textbf{Step 2}  For each evaluation method socre $E$, LLM's score is the average of its scores $ \textbf{A} = [A_{1}, \ldots, A_{m}]$ for each ability dimension:
\begin{equation}
E = \frac{1}{m} \sum_{j=1}^{n} A_j  
\end{equation}

\textbf{Step 3} LLM's scores $\textbf{E} = [E_{1},E_{2},E_{3}]$ for each evaluation method are standardized and then averaged to obtain LLM's final score on ZhuJiu:
\begin{equation}
E_{\text{norm}} = \frac{{E_{k} - E_{\text{min}}}}{{E_{\text{max}} - E_{\text{min}}}}
\end{equation}

\section{Evaluation Perspective for Knowledge Ability}
\label{sec:appendix knowledge}
In the evaluation process of knowledge ability, we mainly evaluate from the properties of accuracy, robustness, completeness and timeliness. For each property, we will randomly generate one hundred sets of evaluation data for evaluation. Here we Need to explain the specific indicators of each evaluation.

\begin{itemize}
    \item \textbf{Accuracy}:Evaluate whether the content of the model’s reply is correct through Exact Match(EM) and ChatGPT\citep{openai202chatgpt}, and calculate the accuracy rate in the 100 questions answered correctly by the model.
    \item \textbf{Robustness}:We use the same set of data to use ChatGPT to randomly generate five different ways of asking questions, and then score according to whether the model is stable in replying to different questions generate by the same set of data. The principle of scoring is that the more stable the content of the reply, the higher the score.
    \item \textbf{Completeness}:Integrity: Only for the evaluation of world knowledge, scoring is based on the proportion of standard answers cover in the model's reply content. For example, according to the calculation of a question with a full score of 10, for the data 
\begin{CJK}{UTF8}{gbsn}
“(中国四大发明—包括—火药,指南针,造纸术,印刷术)”
\end{CJK}
    \textit{``(The Four Great Inventions of ancient China—include—gunpowder, compass, P
    papermaking, printing)''}generate the evaluation question 
\begin{CJK}{UTF8}{gbsn}
“中国的四大发明包括哪些?”
\end{CJK} 
    \textit{``What are the Four Great Inventions of ancient China?''}, if the model answers 
\begin{CJK}{UTF8}{gbsn}
“火药,指南针,造纸术,印刷术”
\end{CJK} 
    \textit{``gunpowder, compass, papermaking, printing''}, it will get a full score of 10, and answer 
\begin{CJK}{UTF8}{gbsn}
“火药,指南针,造纸术,瓷器”
\end{CJK} 
    \textit{``gunpowder, compass, papermaking, china''}has a correct rate of 75 percent and a score of 7.5.
    \item \textbf{Timeliness}:It is only aim at the evaluation of world knowledge, and specifically evaluates the update degree of LLMs knowledge, similar to accuracy, and evaluates whether the answer of the model is correct or not according to EM and ChatGPT.
\end{itemize}


\begin{figure}[t]
\centering
\includegraphics[width=\linewidth]{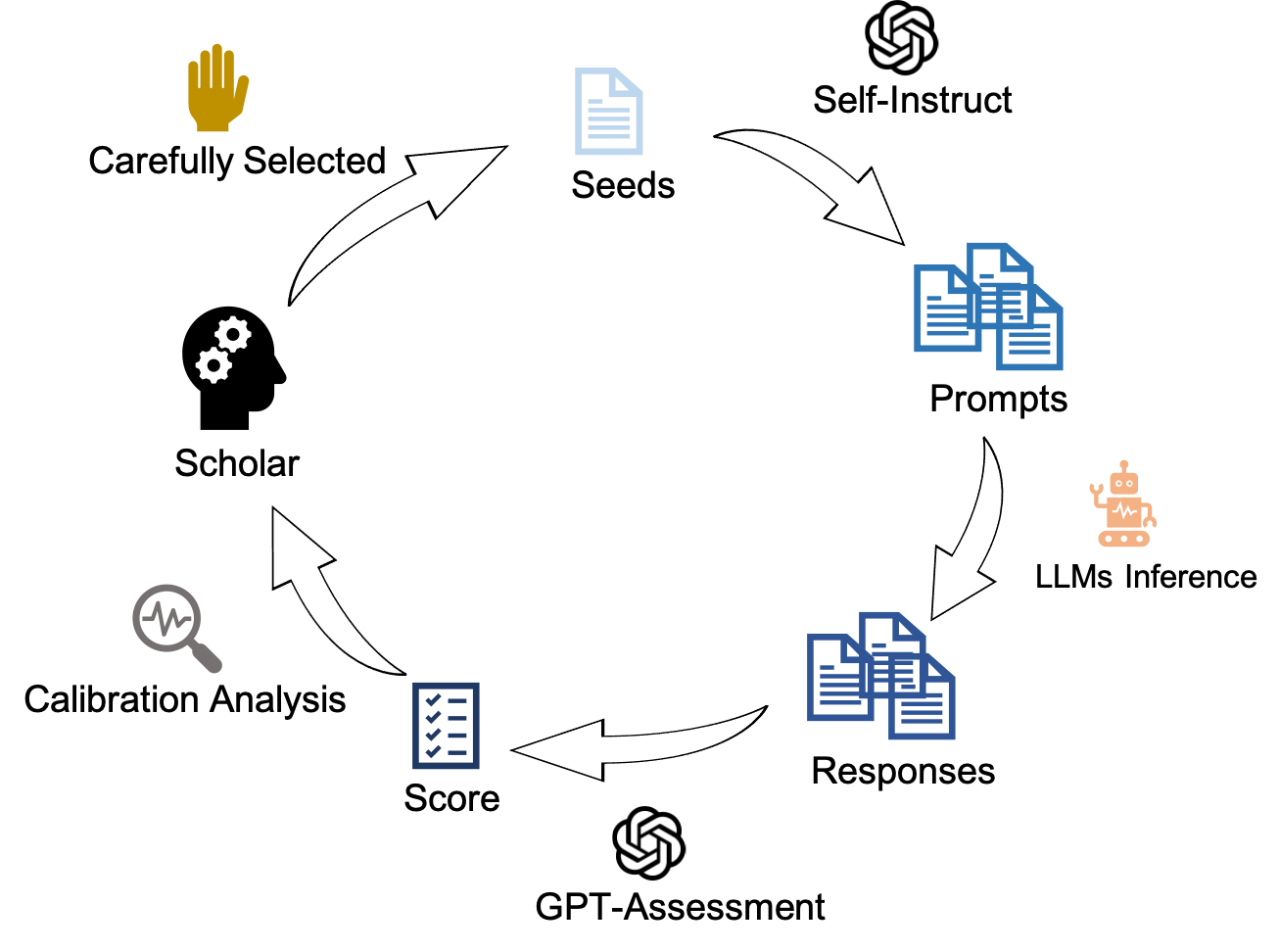}
\caption{The specific processes of data construction and Scoring Evaluation}
\label{fig:figure3}
\end{figure}

\begin{figure*}[t]
    \centering
    \includegraphics[scale=0.5]{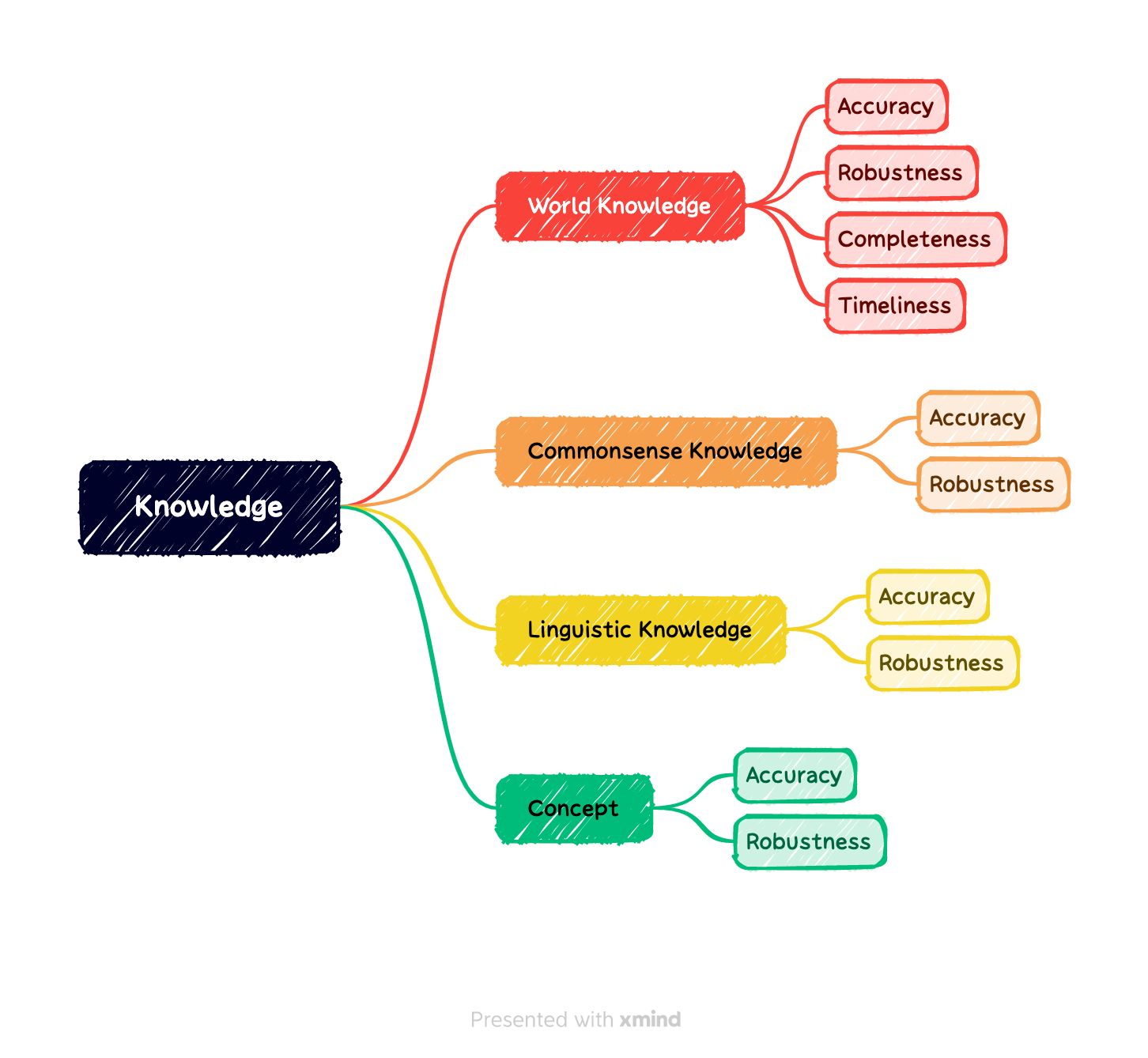}
    \caption{The overall framework of Knowledge benchmark}
    \label{fig:figure4}
\end{figure*}


\begin{figure*}[t]
    \centering
    \includegraphics[width=\textwidth]{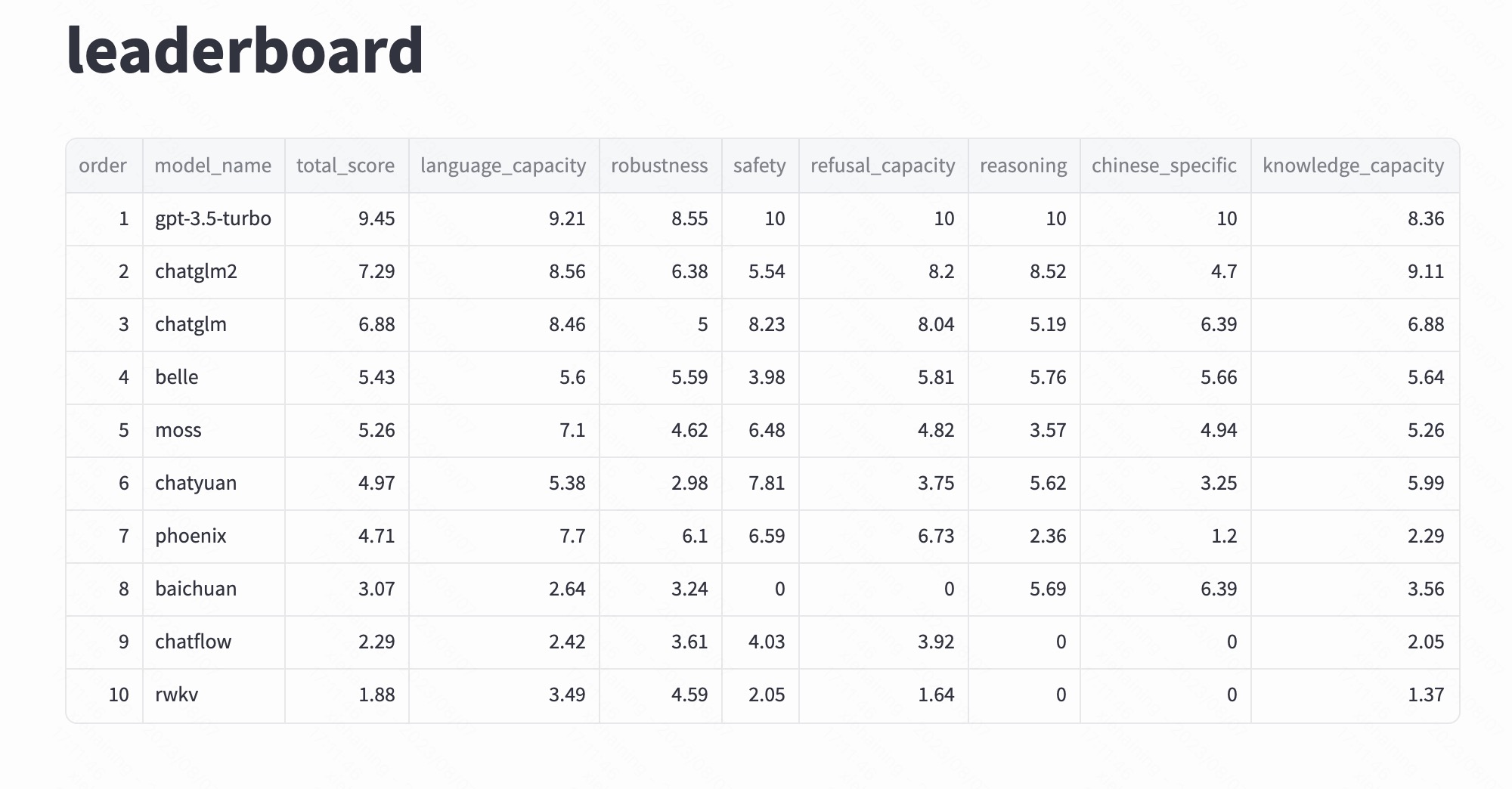}
    \caption{Visualizations of evaluation results}
    \label{fig:figure5}
\end{figure*}

\begin{figure*}[p]
    \centering
    \includegraphics[width=\textwidth]{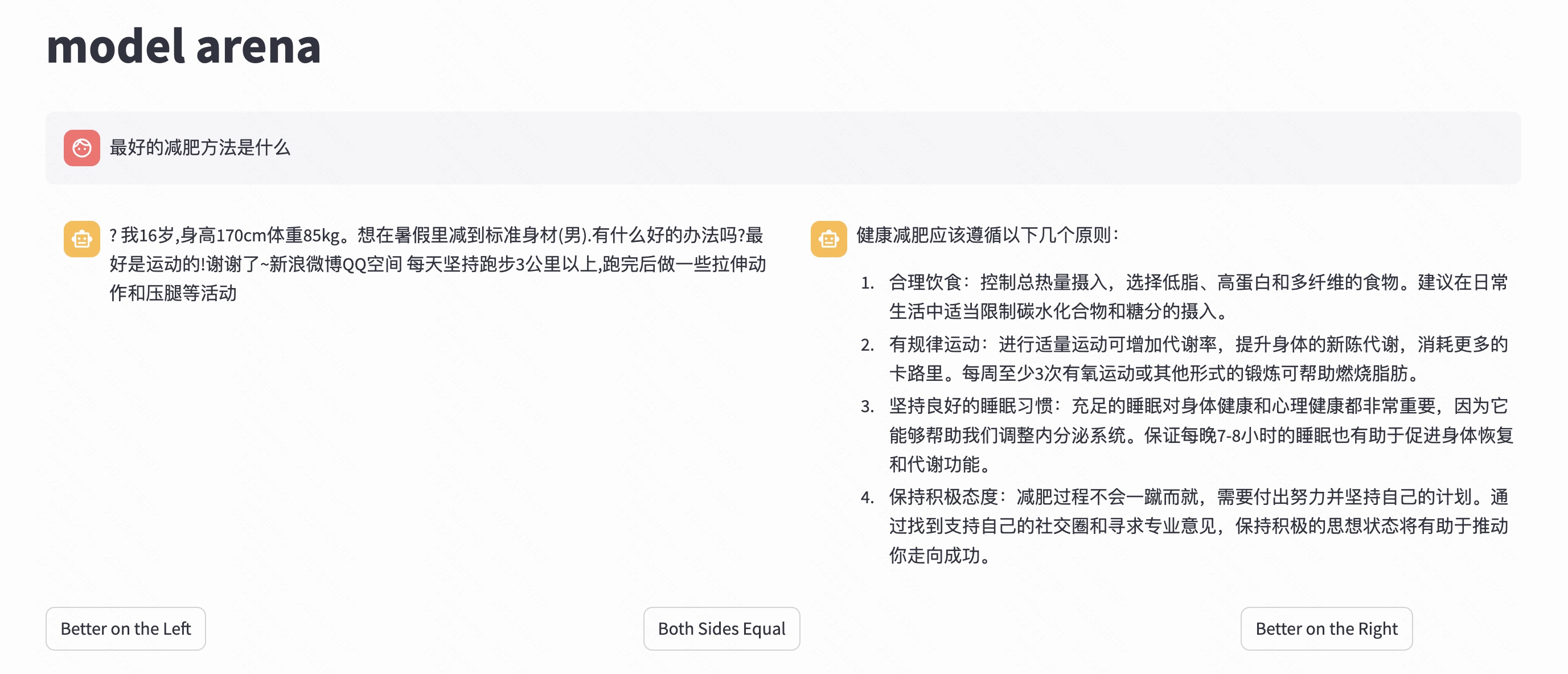}
    \caption{Visualization of Model Arena. And we shown example in English in table \ref{fig:figure7}}
    \label{fig:figure6}
\end{figure*}



\begin{figure*}[p]
    \centering
    \includegraphics[width=\textwidth]{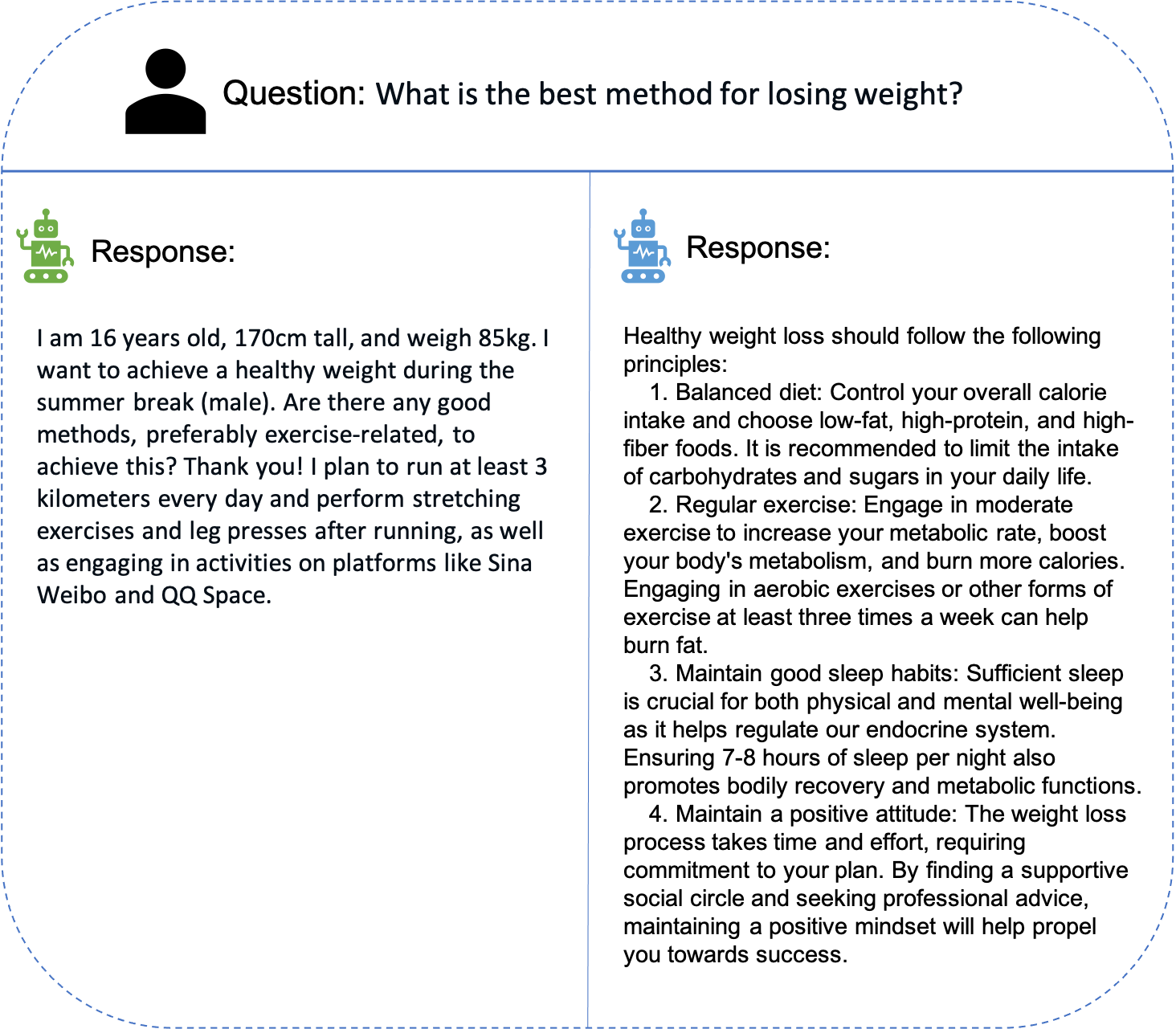}
    \caption{English translation of Model Arena example}
    \label{fig:figure7}
\end{figure*}

\end{document}